\documentclass[conference]{IEEEtran}
\IEEEoverridecommandlockouts

\usepackage{cite}
\usepackage[most]{tcolorbox}
\usepackage{amsmath,amssymb,amsfonts}
\usepackage{algorithmic}
\usepackage{graphicx}
\usepackage{textcomp}
\usepackage{hyperref}   
\usepackage{hyphenat}   
\usepackage{xcolor}
\usepackage[linesnumbered,ruled,vlined]{algorithm2e}
\usepackage{url}
\usepackage{xcolor}
\definecolor{grey}{rgb}{0.5,0.5,0.5}
\def\BibTeX{{\rm B\kern-.05em{\sc i\kern-.025em b}\kern-.08em
    T\kern-.1667em\lower.7ex\hbox{E}\kern-.125emX}}
\begin{document}

\title{A Framework for Analyzing Abnormal Emergence in Service Ecosystems Through LLM-based Agent Intention Mining}

\author{
\IEEEauthorblockN{Yifan Shen$^{1,2,3,\dagger}$, Zihan Zhao$^{1,2,3,\dagger}$, Xiao Xue$^{1,2,3,*}$, Yuwei Guo$^{1,2,3}$, Qun Ma$^{1,2,3}$, Deyu Zhou$^{4,5}$, Ming Zhang$^{6}$}\\
\IEEEauthorblockA{$^1$\textit{College of Intelligence and Computing}, \textit{Tianjin University}, Tianjin, China \\
$^2$\textit{Tianjin Key Laboratory of Healthy Habitat and Smart Technology}, Tianjin, China \\
$^3$\textit{Laboratory of Computation and Analytics of Complex Management Systems}, \textit{Tianjin University}, Tianjin, China \\
$^4$\textit{School of Software}, \textit{Shandong University}, Jinan, China \\
$^5$\textit{Joint SDU-NTU Centre for Artificial Intelligence Research (C-FAIR)}, \textit{Shandong University}, Jinan, China \\
$^6$\textit{Faculty of Environment, Science and Economy}, \textit{Exeter University}, Exeter, United Kingdom \\\\
Email:  \{shenyifan0910, zhaozihan, jzxuexiao, 2024244171,1023244018\}@tju.edu.cn, \\
zhoudeyu@mail.sdu.edu.cn, zhangming1015518539@outlook.com}
\thanks{* Corresponding author: Xiao Xue (\textit{jzxuexiao@tju.edu.cn})}
\thanks{\textsuperscript{\dag} These authors contributed equally to this work.}
}

\maketitle
\begin{abstract}
With the rise of service computing, cloud computing, and IoT, service ecosystems are becoming increasingly complex. The intricate interactions among intelligent agents make abnormal emergence analysis challenging, as traditional causal methods focus on individual trajectories. Large language models offer new possibilities for Agent-Based Modeling (ABM) through Chain-of-Thought (CoT) reasoning to reveal agent intentions. However, existing approaches remain limited to microscopic and static analysis.
This paper introduces a framework: Emergence Analysis based on Multi-Agent Intention (EAMI), which enables dynamic and interpretable emergence analysis. EAMI first employs a dual-perspective thought track mechanism, where an Inspector Agent and an Analysis Agent extract agent intentions under bounded and perfect rationality. Then, k-means clustering identifies phase transition points in group intentions, followed by a Intention Temporal Emergence diagram for dynamic analysis. The experiments validate EAMI in complex online-to-offline (O2O) service system and the Stanford AI Town experiment, with ablation studies confirming its effectiveness, generalizability, and efficiency. This framework provides a novel paradigm for abnormal emergence and causal analysis in service ecosystems. The code is available at \url{https://anonymous.4open.science/r/EAMI-B085}.
\end{abstract}


\begin{figure}[t]
    \centering
    \includegraphics[width=\columnwidth]{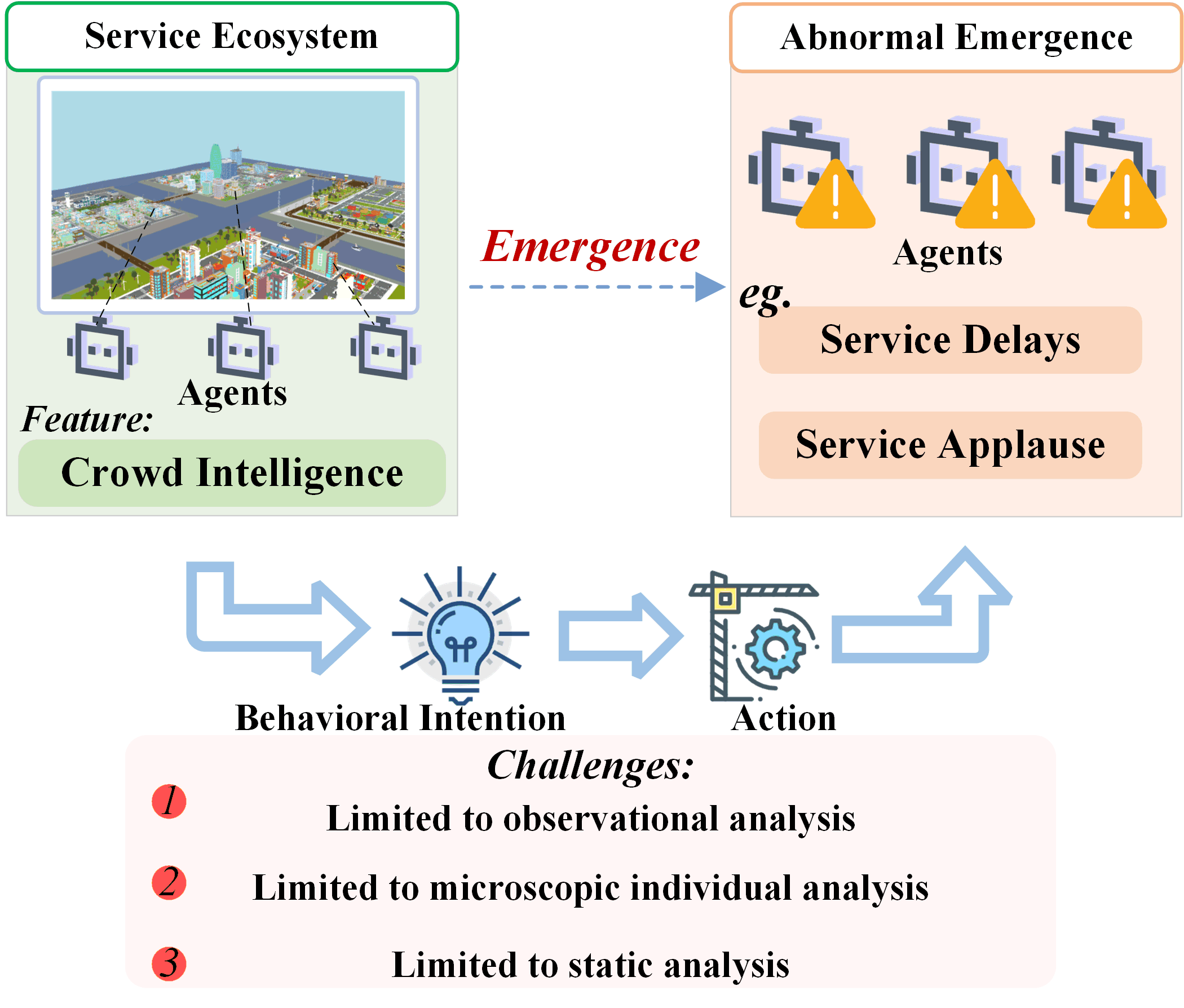}
    \caption{Research background of emergence analysis in service ecosystems.}
    \label{fig:fig1}
\end{figure}

\section{Introduction}
With the rapid advancement of emerging information technologies, such as service-oriented computing, cloud computing, the Internet of Things (IoT), and blockchain, an increasing number of enterprises and organizations are undergoing a service-oriented transformation. They encapsulate their business capabilities, including applications, platforms, data, algorithms, and resources, into diverse services. These services, which can be Web Services, O2O services, OpenAPIs, or Mobile APPs, are dynamically combined and collaborated through technologies such as service composition / mashup, workflow management, and service customization. This process enables cross-border integration on the Internet. Within this technical framework, the entire service landscape is redefined by the logic of service. The service system evolves into a service ecosystem, which is jointly operated by numerous intelligent service agents \cite{xue2019analysis, xue2023chatgpt, xiao2023putational}. Agents refer to entities within the service ecosystem capable of autonomously perceiving environmental information, making decisions, and taking actions to achieve specific service objectives. These agents, including individuals, enterprises, or intelligent robots, collaborate to drive the continuous iterative evolution and self-growth of the service ecosystem. Their cooperation endows the service ecosystem with extraordinary energy and vitality\cite{xue2022research,kang2022ai,xue2024computational, xue2023computational}.

The competition-cooperation dynamics among service ecosystem agents generate emergent phenomena through causal interactions. In intelligent customer service systems, individual responses may induce non-linear information cascades causing operational delays, necessitating causal analysis between micro-level behaviors and macro-outcomes \cite{holland2000emergence,bikhchandani1992theory}. Conventional methods focusing on observable actions ignore intentional states as causal drivers \cite{nagin2001analyzing}. Although modern agents based on the Large Language Model (LLM) using Chain-of-Thought (CoT) reasoning \cite{wei2022chain} enhance decision transparency, their static single agent architectures lack the temporal progression and multi-agent interactions required to explain abnormal emergence as shown in Fig \ref{fig:fig1}, analyzing abnormal emergence phenomena in service ecosystems faces the following challenges:

\begin{itemize}
  \item Limited to observational analysis: Traditional causal analysis methods only make analytical inferences by observing the external behaviors of agents (such as action frequency and interaction objects) and are unable to analyze cognitive reasons such as goal conflicts and flaws in reasoning logic behind their decision making.
  \item Limited to microscopic individual analysis: Emerging LLM-based agents provide the possibility to observe the intentions behind the behaviors of agents. However, CoT technology only focuses on the analysis of individual agents' thoughts, ignoring the transformation and dissemination of thoughts into intentions among groups.
  \item Limited to static analysis: Existing research methods achieve causal reasoning based on the final results of the service system, which is a static process. The abnormal emergence of complex systems occurs in the vertical causality, that is, in the causal relationship in the process of time evolution. Therefore, it is crucial to perform dynamic analysis of abnormal emergence considering time evolution.

\end{itemize}

Some existing work has attempted to solve the above-mentioned problems. Yang et al.\cite{yang2025finding} used reinforcement learning to determine the occurrence of causal emergence. However, this study failed to break away from the dilemma of observational analysis and did not further delve into thought or intention level of agents. Park et al.\cite{park2023generative} proposed the concept of generative agents and used LLM-based agents to simulate an AI town to observe emergence. Although emergence phenomena such as election events were observed in the experiment, their research did not conduct an in-depth analysis of the thinking changes of the agent group or attempt to explain the emergence phenomena.  

To address these challenges, this paper introduces a framework: Emergence Analysis based on Multi-Agent Intention (EAMI), which links the intention of microscopic agents with macroscopic service emergence. In the EAMI framework, the Inspector Agent first tracks the thoughts of each agent in the ABM. Subsequently, the Analysis Agent will examine the changes of each agent to analyze whether new intentions are generated. Then, word embedding and clustering are carried out on the newly emerged intentions. Finally, a Intention Temporal Evolution diagram is obtained, and the whole process of analysis from the behavioral intentions of microscopic individuals to macroscopic abnormal emergence phenomena is completed. Our main contributions are as follows.   

\begin{itemize}
  \item \textbf{Dual-Perspective intention Analysis}: We have designed a dual-perspective thought track mechanism to simultaneously capture the rational analysis and intuitive responses in an agent's decision-making. This mechanism integrates Inspector Agent and Analysis agent to monitor the thoughts of multiple agents and detect the emergence of new intention within the group.
  \item \textbf{Dynamic Emergence Analysis of Multi-Agents}: By applying natural language embedding and clustering techniques to the group intentions in a multi-agent system, we have presented a Though Temporal Evolution diagram of multi-agents. This diagram builds a bridge between the microscopic behaviors of individuals and the macroscopic emergence at the system level, and provides a new method for analyzing the abnormal emergence phenomena of the service ecosystem.
  \item \textbf{Comprehensive Experiments}: Through EAMI verification experiments in scenarios such as resource competition in the O2O service system and the Stanford AI Town scenario, we have demonstrated the ability of this framework to analyze the emergence phenomena of the service ecosystem, as well as its universality in different application scenarios.
\end{itemize}

\section{Background and Motivation}

In the research of the service ecosystem, understanding the intention and behaviors of agents is crucial. This section will systematically review the emergent phenomena in service ecosystem, as well as the thoutht construction methods based on LLMs, and introduce the relevant research works related to them and the existing problems.

\subsection{Emergence Phenomena in Service Ecosystem}
With the advent of the new era of intelligent interconnection of all things, a large number of service systems in which numerous intelligent agents collaborate through division of labor have emerged in the field of modern service industry\cite{vargo2017conceptualizing}. A representative example is Huawei's HarmonyOS service ecosystem, which enables cross-device intelligence by coordinating interactions among users, services, and environments. Such ecosystems typically feature heterogeneous agents with varying intelligence levels, complex interdependencies, and dynamically evolving architectures. The emergence phenomenon refers to the macroscopic manifestation of the system as a whole that exceeds the capabilities of individual intelligent agents\cite{chandler2019does}. Peters et al. \cite{peters2016heteropathic}believe that heteropathic resource integration may lead to new emergent properties in service ecosystems. Qian et al.\cite{qian2024scaling}have studied the causes of emergence phenomena in multi-agent collaboration networks by means of topology, and Yang et al. \cite{yang2025finding}have used reinforcement learning to determine the occurrence of causal emergence. However, existing research still has deficiencies. It mainly focuses on the behaviors of agents, ignoring the intentions behind their behaviors. Further research is still needed for an in-depth understanding and analysis of the emergence phenomena in service ecosystems.

\subsection{Thoughts of LLM-based Agents}
\begin{figure*}[t] 
    \centering
    \includegraphics[width=\textwidth]{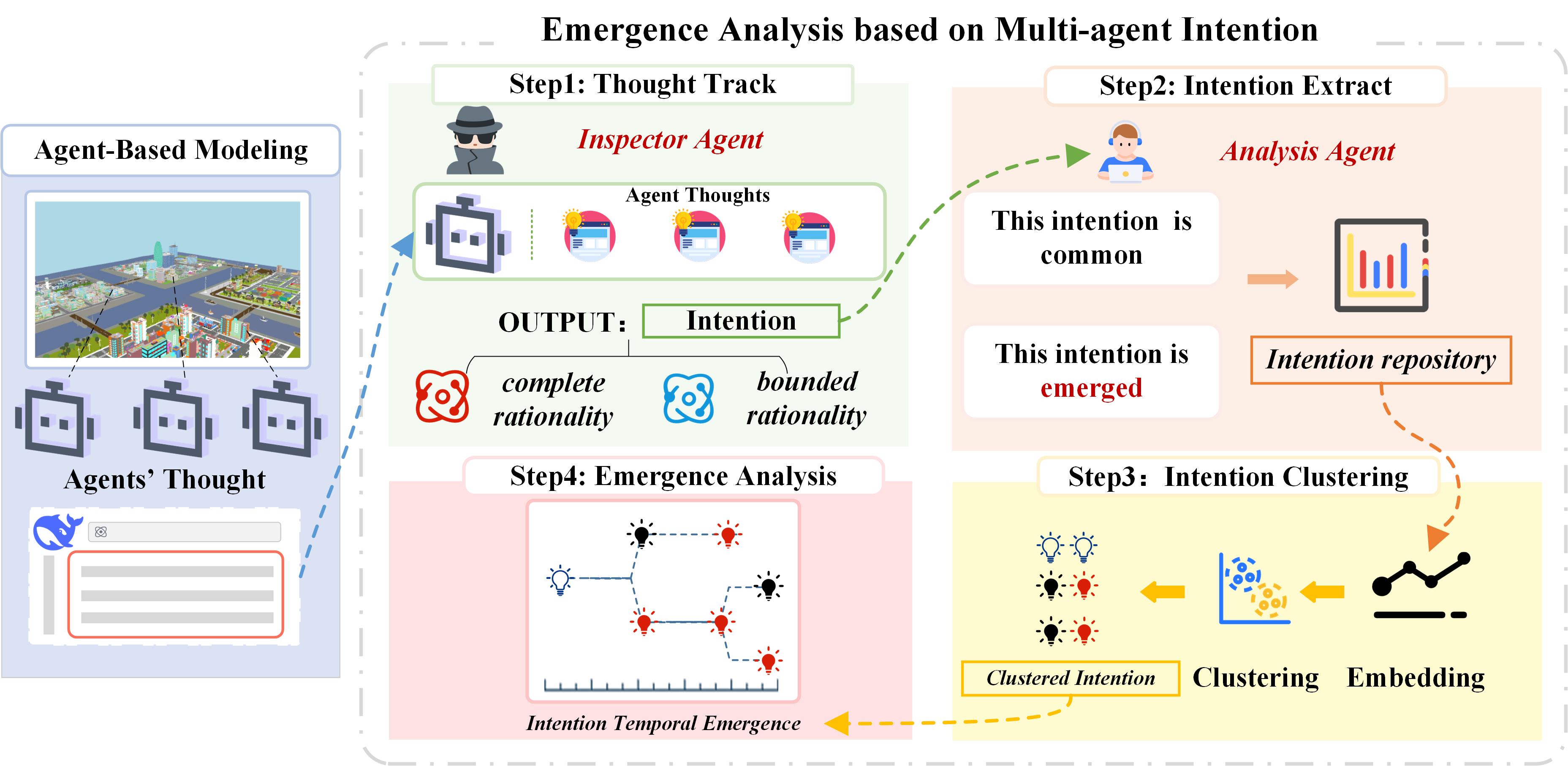}
    \caption{The EAMI (Emergence Analysis based on Multi-Agent Intention) framework, including Individual-Thought-Track, Emergent-Intention-Extract, Group-Intention-Clustering and System-Emergence-Analysis.}
    \label{fig:fig2}
\end{figure*}

The research work related to the thought of LLM-based agent covers multiple aspects. Currently popular prompt engineering frameworks and reasoning techniques provide support for the thinking of agents, such as methods like ReAct\cite{yao2023react}, CoT\cite{wei2022chain}, and Tree-of-Thought (ToT)\cite{yao2024tree}. These methods enhance the thinking and reasoning abilities of agents. In addition, DeepSeek-R1 \cite{deepseekai2025deepseekr1incentivizingreasoningcapability} enhances the reasoning ability of LLM through reinforcement learning, and is able to achieve powerful CoT reasoning with minimal complexity. In a multi-agent environment, the "Hypothetical Minds" model proposed by Stanford University combines LLM and multi-agent reinforcement learning\cite{cross2024hypothetical}. By generating, evaluating, and refining hypotheses about the strategies of other agents, it improves the performance of agents. The development of these research findings and models provides new ideas and methods for the application of agents in complex tasks. Although these methods effectively improve the reasoning and decision-making abilities of agents, they are still limited to single-agent scenarios. For example, the Graph-of-Thought (GoT)\cite{besta2024graph} method optimizes task solving by constructing a thought graph, but it does not analyze the reasons for the emergence of thought among agents.

\subsection{Motivation}
Currently, the analysis methods for emergent phenomena in service ecosystem are mainly limited to behavioral observation, while neglecting intention analysis. In terms of constructing the thoughts of LLM based agents, existing methods are mainly CoT and its improved versions, such as CoT-SC\cite{wang2022self}, Tree-of-Thought (ToT), Graph-of-Thought (GoT), etc. Although these methods have laid the foundation for the construction of agent thoughts, they mainly focus on the individual agent level and have not delved deep into the system level to utilize agent intention for analyzing and explaining the emergent behaviors of service systems. Therefore, we propose the EAMI framework. Starting from the individual thoughts of multiple agents, it conducts in-depth analysis to obtain the intentions of agents, and then proceeds with the analysis of the dynamic evolution process from the system level.

\section{Methodology}

As shown in Fig \ref{fig:fig2}, the EAMI framework is a hierarchical architecture that bridges microscopic agent intention to macroscopic service emergence through four tightly coupled steps: \textit{Individual Thought Track}, \textit{Emergent Intention Extract}, \textit{Group Intention Clustering}, and \textit{System Emergence Analysis}. Fig ~\ref{fig:fig2} illustrates the workflow. In the following, we formalize each module with mathematical rigor and implementation specifics.

\subsection{Individual Thought Track}

Tracking and obtaining the thinking process of agents during the simulation is a prerequisite for subsequent emergence analysis. However, tracking the thoughts of agents is a challenging task. In this section, we introduce and implement the Inspector Agent, whose purpose is to track and record the entire thinking process of each agent during the simulation.

\paragraph{Inspector Agent}In order to track thoughts of agents, we introduce an \textit{Inspector Agent} which is responsible for monitoring and extracting the thinking process of all agents in the system. For each agent $A_i$, the Inspector Agent obtains thoughts of the agent from two perspectives: \textit{bounded rationality} (instinct-driven) and \textit{complete rationality} (goal-oriented). This dual-perspective approach to obtaining the agents' thoughts enables us to comprehensively understand the decision-making process at the individual level.

\paragraph{Dual-Perspective Thought Extraction}
Given a decision-making situation $q$, the following formula demonstrates the process by which the Inspector Agent generates parallel thinking streams for $A_i$:

\begin{equation}
\begin{aligned}
c_s^{(i)} & = \text{LLM}_{\phi_s}\left(q; \mathcal{M}^{i}, T^{i}\right) \quad  \\
c_r^{(i)} & = \text{LLM}_{\phi_r}\left(q; \mathcal{M}^{i}, T^{i}\right) \quad
\end{aligned}
\end{equation}

where $c_s^{(i)}$ represents the thought of bounded rationality, $c_r^{(i)}$ represents the thought of complete rationality, $\mathcal{M}^{i}$ is the memory of $A_i$ which includes the integration of past experiences and information from environmental interactions. $T^{i}$ is the current thinking state, which here is the contextual information. ${LLM}_{\phi}$ represents the process by which the Inspector Agent extracts thought. This process is implemented by calling the LLM interface. The LLM mentioned here and in the following parts of this paper all adopt the locally deployed DeepSeek\footnote{\url{https://huggingface.co/deepseek-ai/DeepSeek-R1-Distill-Qwen-32B}} large language model.

Through the above process, we have successfully tracked the agents' thoughts from dual-perspective. The above thoughts will be transformed into intentions and provided for the next step.

\subsection{Emergent Intention Extract}
This section mainly focuses on identifying and extracting key emergent intention during the evolution process of the agents' intentions. To this end, we equip each agent with a dedicated Analysis Agent. The role of Analysis Agent is to analyze the thinking process of the agent and extract the key emergent intention from it. The emergent intention here refers to the novel intention that the agent suddenly exhibits and did not exist before.

\paragraph{Analysis Agent} In the previous step, we obtained the analysis processes \( c_s^{(i)} \) and \( c_r^{(i)} \) of the agent. The Analysis Agent compares them with the stored memory \( \mathcal{M}^{(i)} \) using a function \( g(\cdot) \) and finally determines whether a certain intention is emergent intention. This function \( g(\cdot) \) interfaces with the LLM to detect the emergence of intention, and its return value is a boolean value. The return value of function \( g(\cdot) \) is a boolean value indicating the presence of emergent patterns. When the Analysis Agent determines that the current intention shows sufficient emergence, it returns True and integrates the emergent intention into the group intention repository \( \mathcal{R} \); Otherwise, it returns False, and the repository \( \mathcal{R} \) remains invariant.

Let \( c^{(i)} = (c_s^{(i)}, c_r^{(i)}) \) denote the combined intention of agent \( A_i \). The emergent intention detection is formalized as:
\begin{equation}
\mathcal{R} =
\begin{cases}
\mathcal{R} \cup \ c^{(i)} , & \text{if } g(c_s^{(i)}, c_r^{(i)}, \mathcal{M}^{(i)}) \\
\mathcal{R}, & \text{otherwise,}
\end{cases}
\end{equation}

The above steps conduct an emergence detection on the thinking process, effectively capturing the dynamic evolution of the agent's thinking. By continuously comparing the current intention components with the historical memory, the system can identify and mark important intention changes, and integrate these emergent intentions into the evolving group intention repository.

\subsection{Group Intention Clustering}

In this section, we describe the clustering of intentions \( c \). After intention extraction in the first two steps, the intention \( c \) of each agent is essentially unique. However, directly analyzing these intentions may inadvertently categorize similar intentions as different, resulting in redundant evolutionary paths and complicating the comparison between different intentions. Therefore, it is imperative to systematically classify all \( c \). To do this, we cluster similar intention among the agents to capture the commonalities among them.

After obtaining sentence embeddings using the all-MiniLM-L6-v2 model\footnote{\url{https://huggingface.co/sentence-transformers/all-MiniLM-L6-v2}}, we quantify the semantic similarity between intention representations through cosine similarity. Specifically, the cosine similarity between two intention vectors \( c_i \) and \( c_j \) is computed as follows:
\begin{equation}
    \text{similarity}(c_i, c_j) = \cos(\theta) = \frac{c_i \cdot c_j}{\|c_i\| \, \|c_j\|}
\end{equation}
Here, \( c_i \cdot c_j \) represents the dot product of the vectors, while \( \|c_i\| \) and \( \|c_j\| \) denote their respective L2 norms. This similarity measure allows us to retrieve semantically proximate sentences from the database, ensuring that the most relevant intention expressions are selected.

Subsequently, we apply clustering to the entire set of intention vectors to group similar ideas together. The clustering process is formalized as:
\begin{equation}
    \text{cluster} = k\text{-means}(similarity_{\text{all}}, n_{\text{all}})
    \label{eq:cluster}
\end{equation}
where \( similarity_{\text{all}} \) is the comprehensive similarity matrix computed from all pairwise comparisons of intention vectors, and \( n_{\text{all}} \) denotes the total number of intention points extracted across agents. The \( k\text{-means} \) algorithm \cite{wang2020minilm} is then employed to partition these intentions into clusters, effectively consolidating similar intentions and reducing redundancy. This clustering not only highlights common cognitive themes among agents but also aids in distinguishing unique intention trajectories.

\subsection{System Emergence Analysis}
\begin{algorithm}[t]
\caption{The generation for the Intention Temporal Emergence diagram:}
\KwIn{Clusters $\mathcal{C} = \{C_1, C_2, \ldots, C_n\}$, where each $C_i$ contains agent intention data at time $t_i$}
\KwOut{Intention Temporal Emergence diagram of emergent intentions}

Initialize baseline cluster: $C_0 \gets \varnothing$\;
Initialize a hash table $H$ to store agent influence relationships\;

\For{$i \gets 1$ \KwTo $n$}{
    $\text{EmergentIntentions} \gets C_i \setminus C_{i-1}$\;
    
    \ForEach{intention $t$ in $\text{EmergentIntentions}$}{
        Determine originating agent $a_t$\;
        
        $t_i \gets \text{current time}$\;
        
        $t_{i+1} \gets \text{next time}$\;
        
        $C_i \gets \text{agents at time } t_i$\;
        
        $C_{i+1} \gets \text{agents at time } t_{i+1}$\;
        
        \ForEach{agent $b$ in $C_i \cup C_{i+1}$}{
            \If{agent $b$ is influenced by intentions $t$}{
                Record Intention Temporal Emergence point $(t, a_t, b, t_i)$\;
                
                $H[b] \gets H[b] \cup \{t\}$\;
            }
        }
    }
}
     \Return Intention Temporal Emergence\;
\end{algorithm}
Based on the intention clusters obtained previously, this section will conduct an in-depth analysis of the emergence of intention during the time evolution process of the system. These intention clusters appear at a certain moment and then start to spread among the agents. These characteristics of the intention clusters enable us to accurately determine the moment when a certain emergent intention first appears and comprehensively define its scope of influence. Ultimately, we generate a \textbf{Intention Temporal Emergence Diagram}. The generation algorithm of the Intention Temporal Emergence diagram is shown as Algorithm 1.

The generated Intention Temporal Emergence Diagram stands as an exceptionally powerful analytical tool. It overcomes the limitations of conventional macroscopic observation methods and the challenges in monitoring processual intentions by precisely delineating the critical temporal nodes at which specific intentions first emerge, while meticulously tracking the subsequent impact these intentions have on other agents. In essence, the diagram systematically encapsulates the progression from individual behavioral intentions to collective service emergence, thereby offering a novel conceptual framework for understanding system emergence.

\section{Experiment}
In this section, we conduct extensive experiments to answer the following research questions:

\begin{itemize}
    \item \textbf{RQ1}: Can EAMI explain the emergent phenomena in service simulation?
    \item \textbf{RQ2}: How do the Inspector Agent and Analysis Agent in EAMI affect the emergence analysis?
    \item \textbf{RQ3}: Is EAMI universally applicable to different scenarios when explaining the emergence in service simulations?
\end{itemize}

\subsection{Experimental Setting}
\paragraph{Experiment System}
With the development of mobile networks, the O2O platforms such as Meituan have driven the digital transformation of multiple industries through the integration of online and offline services.\cite{zhao2021online} However, with the rapid growth of these platforms, delivery riders are facing a severe phenomenon of \textbf{involution}\cite{kang2020review}. In order to secure more orders and income, riders are often forced to operate under high levels of work pressure, leading to an increase in physical and mental health burdens and a widening income gap. Additionally, the platform's incentive mechanisms and evaluation systems have further exacerbated the involution. Therefore, to validate the effectiveness of the intention emergence analysis approach, we have constructed a simulation of a real-world multi-agent O2O platform based on LLMs, using this method to analyze the causes and emergence process of the involution phenomenon among delivery riders.
\paragraph{Involution Phenomenon}
The concept of "Involution" was put forward by sociologist Alexander Gerschenkron\cite{gerschenkron2015economic} and is used to describe the state of stagnation or degradation in socio-economic development. In the service ecosystem, involution refers to the excessive competition among individuals or groups in an environment with limited resources, resulting in a decrease in the input-output efficiency and a decline in the overall welfare. In this study, the involution of riders is manifested as follows: They compete for limited orders by extending their working hours and increasing their labor intensity, but their income does not increase significantly. Instead, it leads to health deterioration and income polarization. 
\paragraph{Implementation Details} All experiments are completed using the locally deployed DeepSeek-R1-Distill-Qwen-32B-FP8-Dynamic \footnote{\url{https://huggingface.co/deepseek-ai/DeepSeek-R1-Distill-Qwen-32B}}model, with the temperature parameter set to 0 to ensure reproducibility.

In the experimental environment, we construct a multi-agent system comprising five types of agents: merchants, riders, users, government, and platform. We focus on riders as key objects to simulate the service system, setting up 100 rider agents for the study, each with the aim of minimizing working hours, maximizing order completion, and reducing labor costs. These agents make autonomous decisions, such as selecting working hours and accepting orders, based on LLMs. The simulation runs for 3,600 steps, representing a one-month cycle, with the physical environment modeled as a grid map. Detailed experimental parameters are provided in the appendix.
\paragraph{Dataset Comparison}

\begin{figure}[h]
    \centering
    \includegraphics[width=1\columnwidth]{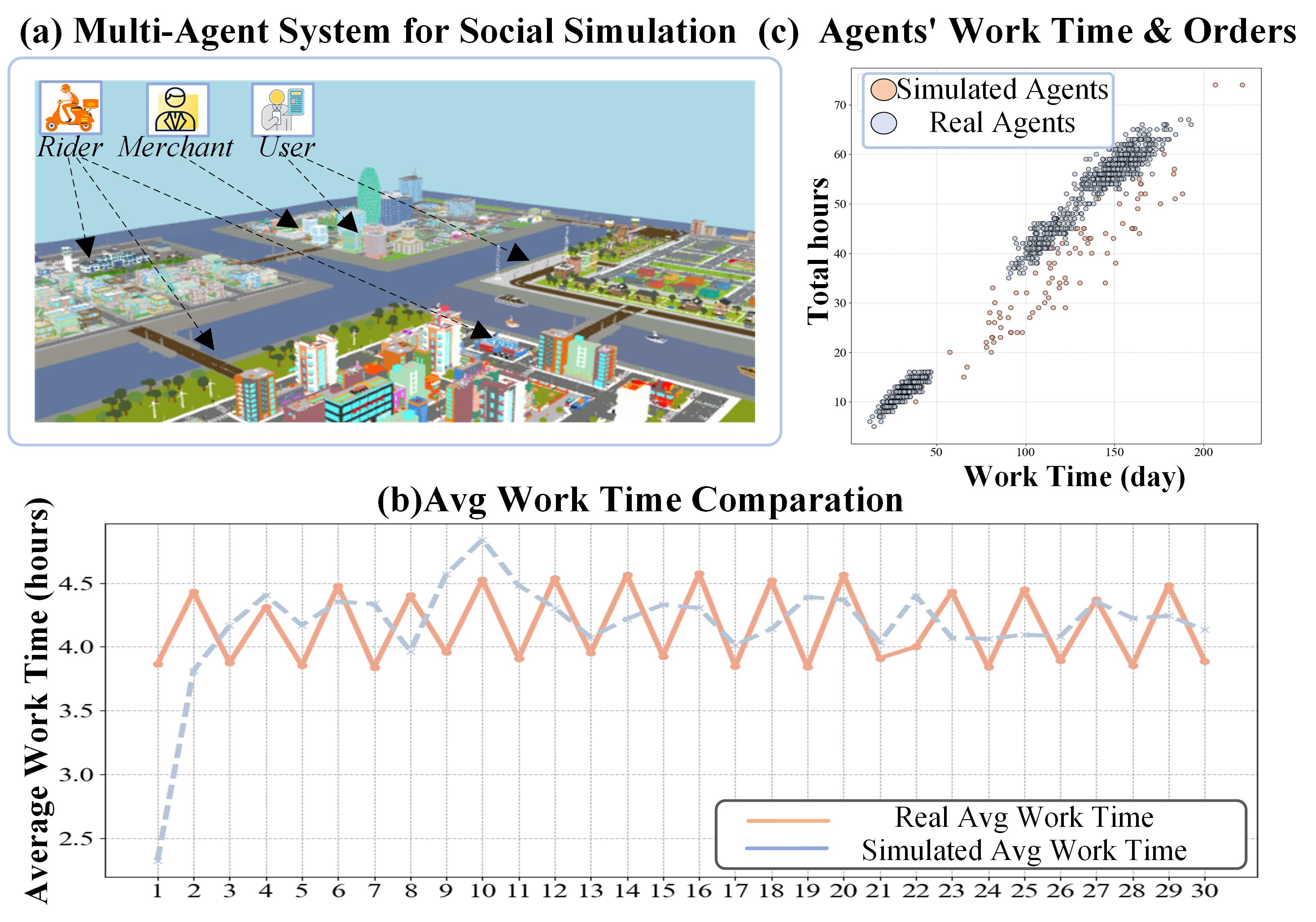}
    \caption{Comparison of the Multi-Agent O2O service system simulation system with real-world datasets.}
    \label{fig:fig4-1}
\end{figure}

\begin{figure*}[t] 
    \centering
    \includegraphics[width=\textwidth]{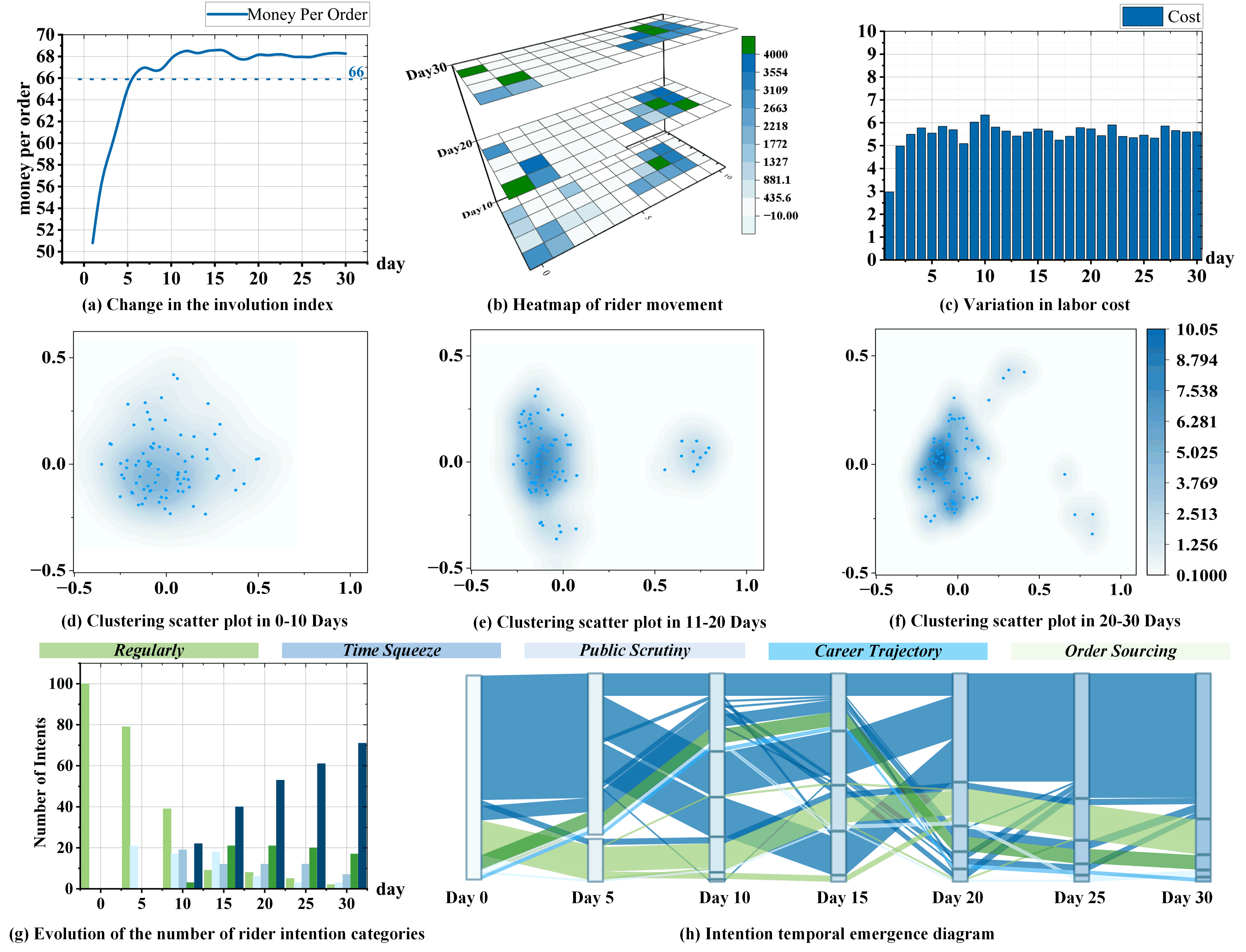}
    \caption{The experimental results are summarized in several figures: Fig (a) illustrates the involution index; Fig (b) and (c) present traditional observational analyses; Fig (d)–(f) display clustering scatter plots of rider intentions at 10-day intervals over a 0–30 day period; Figure (g-h) shows the final intention temporal emergence.}
    \label{fig:fig4-2}
\end{figure*}

To ensure the authenticity and validity of the experiment, we compare real-world data \footnote{\url{https://www.kaggle.com/datasets/saurabhbadole/zomato-delivery-operations-analytics-dataset}} from the Zomato platform \footnote{\url{https://www.zomato.com/}} (including food delivery orders from multiple cities) with our multi-agent system. We mainly compare the average daily effective working hours of riders and the relationship between working hours and order volume, drawing comparisons between the real dataset and the experimental data, as shown in Fig ~\ref{fig:fig4-1}.

Fig ~\ref{fig:fig4-1} (b) shows the average daily effective working hours of riders in both the real dataset and the system simulation. Effective working hours refer to the time spent delivering orders, excluding activities such as waiting. The results indicate a strong alignment between the simulation and real-world data, with riders' working hours fluctuating around 1 hour, confirming the simulation’s validity in modeling real riders' work time decisions.

Fig ~\ref{fig:fig4-1} (c) illustrates the relationship between agents' working hours and the total number of orders accepted. As expected, longer working hours lead to more orders. The grey dots represent 1,320 real riders, while the orange dots correspond to 100 rider agents in the simulation. The close similarity between the simulate and real data validates the experiment’s accuracy and reliability.


\subsection{Simulation Emergence Analysis (RQ1)}

Regarding RQ1, this experiment focuses on the performance of rider agents in the O2O experiment platform. Firstly, we observe the possible phenomenon of involution through the involution index and analyze the behavior trajectories of agents using traditional methods. Subsequently, we explain the emerging phenomena through the EAMI framework, conducting intention clustering analysis and intention emergence analysis in sequence.      

\paragraph{Analysis of the Involution Index} 
Fig \ref{fig:fig4-2}(a) shows the change of the "involution" index within 30 days of the experimental system. By using the labor cost per order as the "involution" index for analysis, it is found that with the passage of time, the labor cost required for a single order shows a significant increasing trend. The labor cost of a single order increases from 50 to 67.5 within 10 days, and the "involution" phenomenon becomes more and more serious after the 10th day.

\paragraph{Analysis of Behavioral Trajectories}
Fig \ref{fig:fig4-2}(b) shows the traditional observation and analysis method. With the help of a heat map, the behavioral trajectories of agents are observed. The heat map intuitively displays the changes in the moving positions of the riders every 10 days. Fig \ref{fig:fig4-2}(c) illustrates the variation in rider labor costs over 30 days, revealing a sudden spike that remains elevated. It can be seen from the figure that initially, the moving positions of the riders are relatively scattered. Every 10 days, the moving positions of the riders become more and more concentrated, the activity range gradually shrinks, and the choice behaviors tend to be consistent. 

\paragraph{Analysis of Intention Clustering}
Fig \ref{fig:fig4-2}(d-f) present the clustering analysis of intentions within 30 days obtained in the first three steps of METEA. Through the analysis framework, five types of intentions are derived. The scatter plots illustrate the distribution of intention points among the five different clusters, with each cluster representing a specific intention. According to the density of the clustering scatter points, it is shown that the intentions of the riders gradually evolve from similar thinking patterns to a variety of thinking patterns. The analysis reveals the emergence of four types of intention clusters, namely: "Going to places with more orders to compete for orders", "Imitation and competition among peers", "Avoiding traffic congestion", "Judgment of order cost-effectiveness", and "Accepting algorithmic allocation".
\paragraph{Analysis of Intention Emergence}
As shown in Fig \ref{fig:fig4-2}(g-h), through the visualized images of the intention evolution process, two main patterns of the riders' thinking changes are demonstrated. Among them, 34 riders believed that they could increase their income by imitating the behaviors of their peers within the first 10 days of the simulation. However, as time went by, they gradually realized that only by going to areas with dense orders to compete for orders could they maximize their income. In addition, 30 riders found that the traffic conditions were poor during the process of regularly accepting orders, so they chose to change the delivery routes to optimize the delivery efficiency. According to the evolution process, it can be concluded that one of the reasons for the emergent involution phenomenon is that riders' pursuit of high-value orders and their pursuit of short-distance transportation are major causes of involution.

The experimental results show that analyzing this process with the aid of the EAMI framework not only further deepens the understanding of the riders' behavior patterns based on the traditional analysis of behavioral trajectories, but also verifies the effectiveness of this framework in analyzing complex service phenomena.

\subsection{Ablation Study(RQ2)}
\begin{figure}[h]
    \centering
    \includegraphics[width=1\columnwidth]{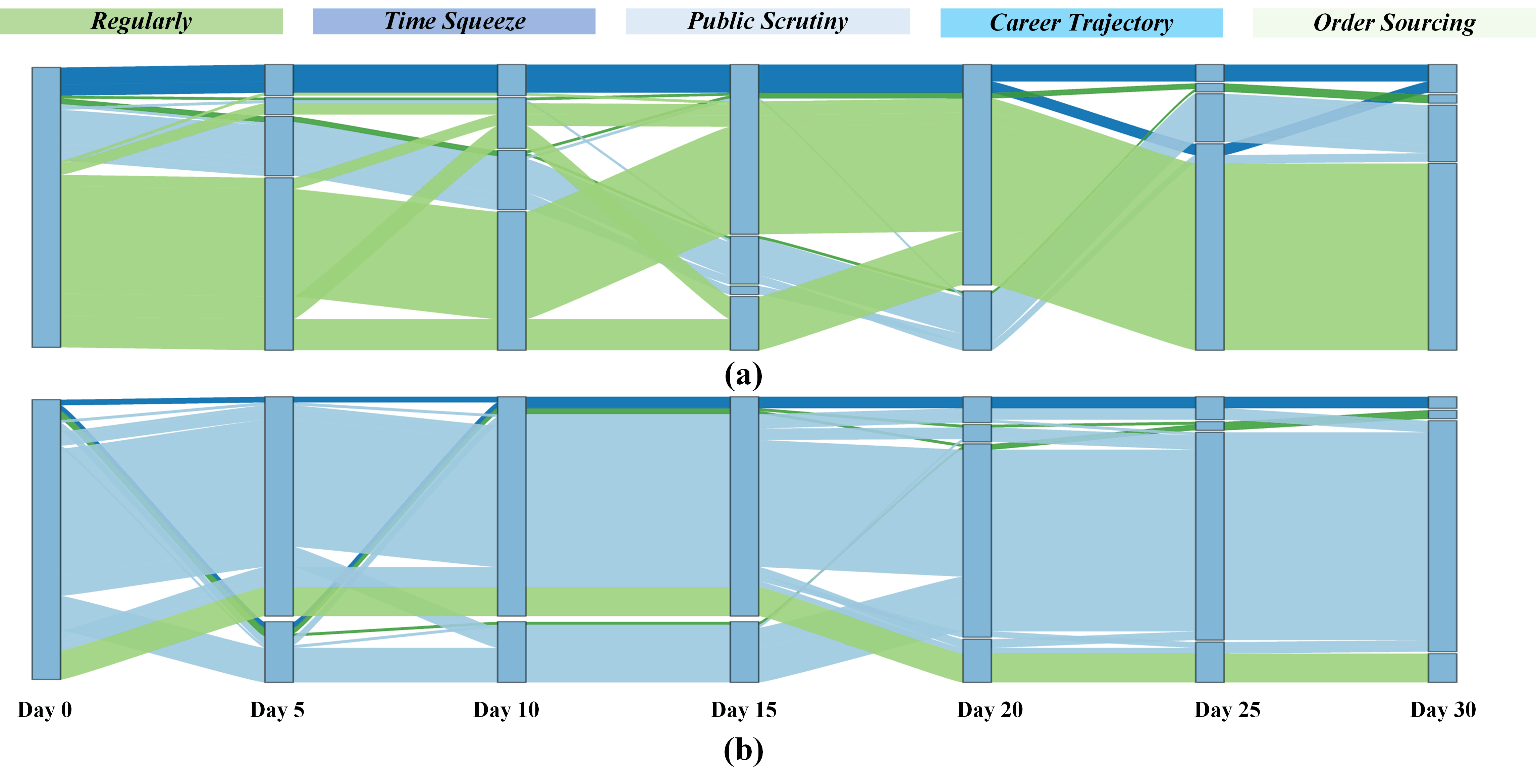}
    \caption{Results of the ablation study on the Individual Thought Track and Emergent Intention Extract modules.}
    \label{fig:fig4-3}
\end{figure}
To verify the functions of each module in the EAMI framework, we conducted ablation experiments, with a focus on evaluating the effectiveness of the Inspector Agent and the Analysis Agent designed in the first two modules. As shown in Fig \ref{fig:fig4-3}(a) after removing the Inspector Agent, the emergent timing diagram fails to capture intentions such as "jealousy" and "imitation and competition among peers". These intentions do not manifest in the CoT process and can only be identified by analyzing them from the perspective of bounded rationality. Fig \ref{fig:fig4-3}(b) After removing the Analysis Agent, the captured information shows that during the period from Day 10 to Day 25, there is a lack of records of intention changes, resulting in the inability to analyze the real-time dynamics of emergent phenomena.

The experimental results demonstrate that the Inspector Agent performs comprehensively in capturing intentions, especially those related to bounded rationality; the Analysis Agent performs precisely in analyzing intention changes. These conclusions collectively prove the high efficiency of the EAMI framework.

\subsection{Universal Scenario Simulation(RQ3)}
To verify whether EAMI has the universality of scenarios when analyzing system emergence, on the basis of a custom O2O service system, we also select the well known Stanford AI Town\footnote{\url{https://github.com/joonspk-research/generative_agents}} as an experimental scenario. We analyze the emergence of agents' intentions in Stanford AI Town through the EAMI framework and attempt to explain the system emergent phenomena that occurred therein.

\begin{figure}[h]
    \centering
    \includegraphics[width=0.9\columnwidth]{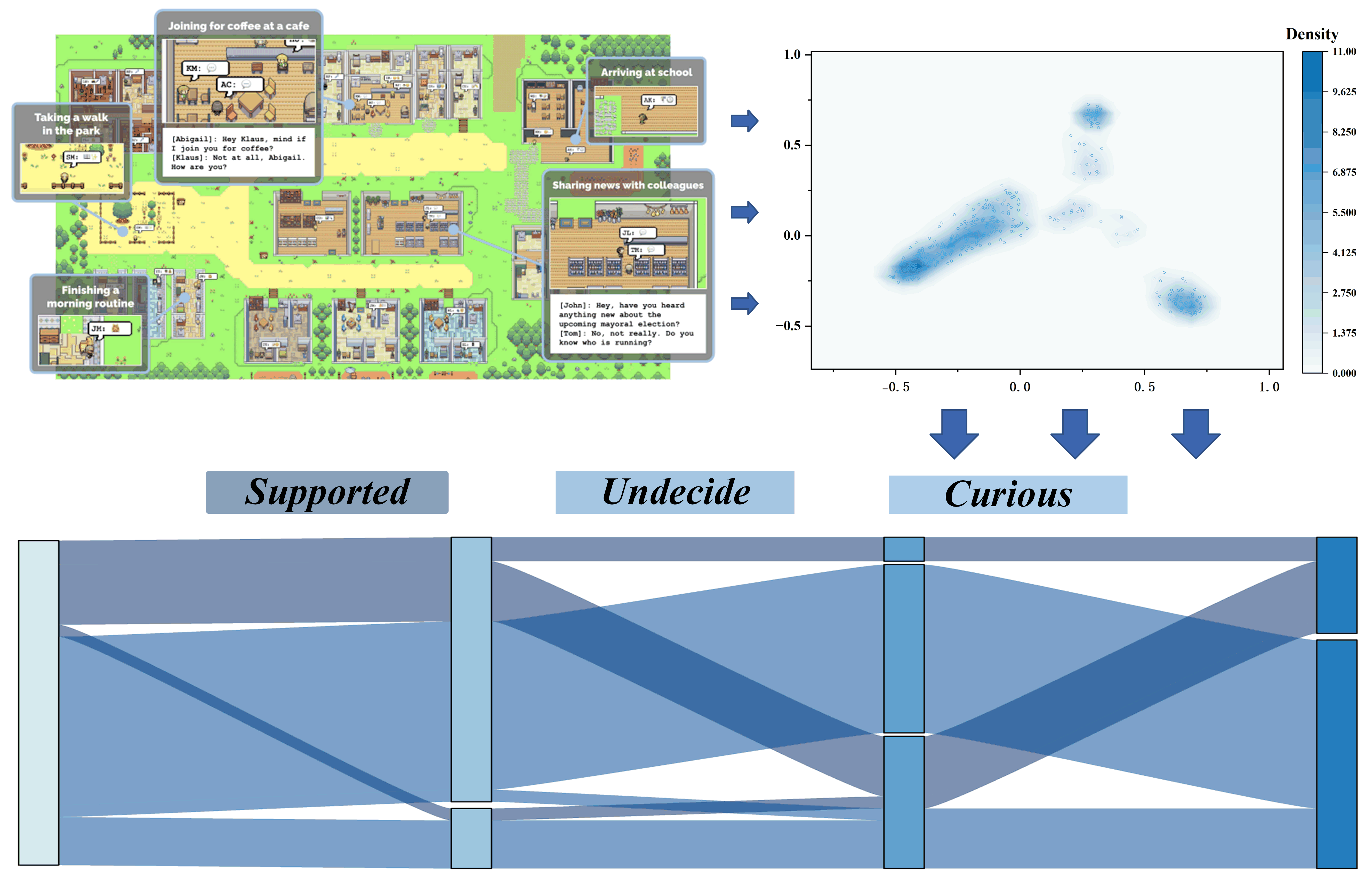}
    \caption{Intention clustering and intention temporal emergence diagram of Stanford AI Town \cite{park2023generative}.}
    \label{fig:fig4-4}
\end{figure}

In Stanford AI Town, one agent expresses its intention to run for election. This event quickly emerge a hot topic in the town, with other agents displaying either supportive or uncertain attitudes toward the election\cite{park2023generative}. As shown in the Fig \ref{fig:fig4-4}, we replicate the experiment and conduct an emergence analysis of the election topics that emerged in the system using the EAMI framework, obtaining the intention clustering and the temporal emergence process of intention. It can be seen that agents initially generally showed a hesitant attitude. However, with the campaign speeches of the candidates, the attitude of agents gradually changed to curiosity, and finally, 70\% of agents formed the intention to support. These intentions spread among the agents, ultimately leading to the emergence of system results.

In conclusion, the EAMI framework effectively explains the emergent election event in the system by analyzing the intentions of multiple agents and uncovering the underlying causes of their behaviors. This validation highlights the universality of the EAMI framework across different scenarios.

\section{Conclusion}
In order to effectively analyze the abnormal emergence phenomena in the service ecosystem, We propose a novel emergence analysis framework based on multi-agent intention, EAMI, which builds a bridge between the individual micro-behaviors and the system-level macro-emergence in LLM-based agent simulation. Leveraging the convenience of LLMs to extract thoughts offers new perspectives on explaining the emergence of complex systems in ABM. EAMI utilizes Inspector Agents and Analysis Agents to track and analyze individual agent intention, employing natural language embedding and clustering techniques to characterize emerging intention. In addition, we construct a highly realistic O2O service system, use EAMI to analyze the emerging involution phenomenon within it, and then conduct an emergence analysis of the Stanford AI Town scenario and ablation experiments. Through these efforts, we effectively verify the effectiveness, universality, and high efficiency of the EAMI framework.

\section*{Limitations}

\paragraph{Simulation Environment} Although our simulation system includes five types of agents: rider agents, merchant agents, customer agents, platform agents, and government agents, only rider agents are powered by LLM. The behavioral logic of the remaining agents is replaced using traditional rule-based methods. Considering the duration of the experiment, the local performance, and the experimental effects in full, the number of rider agents is set to 100, and the simulation duration is one month. If it is possible to run the simulation on a larger scale and over a longer timescale, there will be an opportunity to obtain more representative experimental data.
\paragraph{LLM and Optimization}This paper uses the DeepSeek-R-Distill-Qwen-32B-FP8 -Dynamic model, a distilled LLM with 32 billion parameters. Using the full-fledged DeepSeek R1 (671B parameters) can boost agents' intelligence and behavior. Additionally, fine - tuning the LLM with real-world rider and delivery data is viable, making agents' actions mirror those of actual food-delivery riders more closely.

\bibliography{custom}
\bibliographystyle{IEEEtran}
\cleardoublepage
\appendix
\section{Appendix}
\label{sec:appendix}
\subsection{Agent Descriptions}
In our experimental setup, a total of 100 generative agents were deployed. Each of these agents was assigned the role of a delivery rider within the simulated environment. Agents possess different personality traits and role descriptions. An example of agent description is shown as follows:
\begin{tcolorbox}[colback=grey!20, boxrule=0pt, arc=0pt, sharp corners]
You are Chloe Lewis, a 26-year-old male delivery rider with years of experience behind the wheel. You know the best routes and handle deliveries with care, always looking for ways to optimize your time on the road.
\end{tcolorbox}

\subsection{Prompts}
We employ the LLM-based Inspector Agent to track the agents' thoughts from the dual-perspective of perfect rationality and bounded rationality. The relevant prompts are as follows.
\begin{tcolorbox}[colback=grey!20, boxrule=0pt, arc=0pt, sharp corners]
You should think in a completely rational way, without considering personal characteristics, and you need to do more mathematical calculations and analysis.

You should think in a bounded rational way, mainly considering your personal character without rational calculation and analysis.
\end{tcolorbox}
In the experiment conducted in this paper, the LLM-based agents face two decision-making tasks. One is to decide their work time. The other is to determine whether to accept orders. We sketch prompts for the LLM-based operations of our experiments as follows.
\begin{tcolorbox}[colback=grey!20, boxrule=0pt, arc=0pt, sharp corners]
You are playing a human, your job is a Meituan rider. You need to earn money by working. You live in a town, the town map is presented as coordinates. 

Among 100 riders, your riding distance yesterday ranked 34, the amount of money earned ranked 28, and the number of orders received ranked 31. The time you started working yesterday was 10:00, and the time you finished working yesterday was 18:00, Please use this information to decide whether you want to change your working hours today.

Based on the above information, please give your time to go to work today, time to get off work today.
You should think step by step.

- When you speak, you must use the following format in json:

\{
    
    "go\_to\_work\_time": the time, The time should be in the format *:00.
    
    "get\_off\_work\_time": the time, The time should be in the format *:00.
    
\}
\end{tcolorbox}
\begin{tcolorbox}[colback=grey!20, boxrule=0pt, arc=0pt, sharp corners]
You are playing a human, your job is a Meituan rider. You need to earn money by working. You live in a town, the town map is presented as coordinates.

Now there is the following order information, each order information is represented by a list item, including the pickup and delivery locations for orders and the money that can be obtained. The order information is as follows: \{order\_list\}

Your current location is [28,104]. You can currently accept up to 5 orders.

Based on the above information, please give The list of order numbers you choose to take. The order number you choose must be in the given order list. You should think step by step.
                
- When you speak, you must use the following format in json:

\{
    
    "order\_list": A list of only order\_id that you choose to accept.
    
\}
\end{tcolorbox}
\subsection{Agent Thoughts}
In all experiments of this paper, the DeepSeek model's in-depth thinking function is used to track agents' thoughts, which are wrapped in <think></think> tags in the model response. Here is an example of an agent's thought when deciding working hours.
\begin{tcolorbox}[colback=grey!20, boxrule=0pt, arc=0pt, sharp corners]
\( < \)think\( > \)

Alright, so I'm Ava Johnson, a 50-year-old Meituan rider who's into outdoor activities and looking for an active job. Yesterday was my first day working as a delivery rider, and I got some stats back from the system. I need to figure out whether to adjust my working hours today based on how it went. Let me start by looking at the data from yesterday. I rode for a certain distance, but compared to other riders, I ranked 7th out of 100 in all three categories: riding distance, money earned, and number of orders received. Hmm, that's interesting. So, I'm pretty close to being average or maybe slightly above.

...

\( < \)/think\( > \)   
\end{tcolorbox}
\subsection{Experimental Setup}
We have outlined the experimental settings in Section 4.1 (of the main paper). In the following, we exhibit more details of our experimental parameters.

The experiment lasted for 3600 steps, with 120 steps representing one day, simulating a total duration of 30 days. The number of riders was set to 100, with 3 peaks for order generation. The map size was 200x200, and each agent could move a distance of 30 units per step, with a maximum of 3 orders held by an agent at any given time.

\end{document}